\documentclass[10pt,conference]{IEEEtran}
\usepackage{cite}
\usepackage{amsmath,amssymb,amsfonts}
\usepackage{algorithmic}
\usepackage{graphicx}
\usepackage{textcomp}
\usepackage{xcolor}
\usepackage{flushend} 
\def\BibTeX{{\rm B\kern-.05em{\sc i\kern-.025em b}\kern-.08em
    T\kern-.1667em\lower.7ex\hbox{E}\kern-.125emX}}
\begin{document}

\title{Investigating Markers and Drivers of Gender Bias
in Machine Translations}

\author{\IEEEauthorblockN{Peter J Barclay}
\IEEEauthorblockA{\textit{School of Computing, Engineering, and Built Environment, } \\
\textit{Edinburgh Napier University}\\
Edinburgh, United Kingdom \\
p.barclay@napier.ac.uk}
\and
\IEEEauthorblockN{Ashkan Sami}
\IEEEauthorblockA{\textit{School of Computing, Engineering, and Built Environment, } \\
\textit{Edinburgh Napier University}\\
Edinburgh, United Kingdom \\
a.sami@napier.ac.uk}
}

\maketitle

\begin{abstract}
Implicit gender bias in Large Language Models
(LLMs) is a well-documented problem, and implications of
gender introduced into automatic translations can perpetuate
real-world biases.
However, some LLMs use heuristics or post-processing to mask
such bias, making investigation difficult. Here, we
examine bias in LLMs via back-translation, using the
DeepL translation API to investigate the bias evinced
when repeatedly translating a set of 56 Software Engineering
tasks used in a previous study. Each statement starts with
`she', and is translated first into a `genderless' intermediate
language then back into English; we then examine pronoun-choice
in the back-translated texts. We expand prior research in the following ways: (1) by
comparing results across five intermediate languages, namely
Finnish, Indonesian, Estonian, Turkish and Hungarian; (2) by
proposing a novel metric for assessing the variation in gender
implied in the repeated translations, avoiding the
over-interpretation of individual pronouns, apparent in earlier
work; (3) by investigating sentence features that drive bias; (4)
and by comparing results from three time-lapsed datasets to
establish the reproducibility of the approach. We found that
some languages display similar patterns of pronoun use, falling
into three loose groups, but that patterns vary between groups;
this underlines the need to work with multiple languages. We
also identify the main verb appearing in a sentence as a likely
significant driver of implied gender in the translations. Moreover,
we see a good level of replicability in the results, and establish
that our variation metric proves robust despite an obvious change
in the behaviour of the DeepL translation API during the course
of the study. These results show that the back-translation method
can provide further insights into bias in language models.
\end{abstract}


\section{Introduction}
\label{sec:intro}

With increasing use of machine translation and automatic text
generation, it is important to understand the effects of biases in the
language models underlying these technologies. Biases in the models
may produce biases in the generated text, propagating such
biases with real-world implications.

In this study, we investigate the appearance of implied gender when automatically
translating texts. Here, we restrict our attention to how gender may 
be implied by pronoun selection in translations. 

While Web-based translation
services may provide alternative translations with different
pronouns, or disclaimers regarding gender, this just masks bias in
the underlying model, and is not consistently applied. For example, at the time of writing, Google Translate
provides alternative forms when translating into English, but not
when translating into Swedish. However, when larger
quantities of text are generated using translation APIs, one specific translation must given, and the embedded pronoun choices may
reveal biases in the underlying language model.

Considering the complexities of gender in natural language, 
it is not clear how bias in automatic translations should be interpreted,
and we feel that earlier studies have not taken a sufficiently nuanced approach to their analyses.
We start from the observation that some natural languages have grammatical gender, 
a type of noun class where each noun is assigned to a category such as `masculine’, `feminine’, or `neuter’.
Though not universal, such gender systems exist in many languages, including some of the world's
most used languages such as Spanish, Arabic, and Russian. 
We distinguish natural gender,  based on the biological sex or societal 
gender role of a person,  from grammatical gender, which is purely a 
linguistic feature \cite{corbett_gender_1991}.
Natural gender can still be implied by pronoun choice in languages 
whose gender system is  not based on a masculine/feminine classification 
(\textit{eg}, Swedish),  or with no grammatical gender system for nouns (\textit{eg}, English). 

Grammatical and natural gender are not necessarily aligned; 
in French for example, `masculinité’ (masculinity) is grammatically feminine,  
whereas in Gàidhlig (Scottish Gaelic) `boireannach’ (a woman) is grammatically masculine.  
There is a modicum of evidence that grammatical gender in language can affect 
the worldview of speakers of that language,
such as the results presented by Phillips and Boroditsky \cite{phillips_can_2003}. 
This is an example of the  Sapir-Whorf hypothesis, the idea that language
structures can affect perceptions and cognition;
the term was introduced in 1954 by Hoijer \cite{hoijer_language_1954} 
then further discussed by Koerner \cite{koerner_sapir-whorf_1992}. 
However, this hypothesis is still subject to some controversy, and is beyond the scope of this study.

There may exist complicated relationships between grammatical
gender, natural gender, and language usage. For example, in the French sentence `Je suis prête’ (I am ready),
the speaker indicates that she identifies as female by using a feminine adjectival form, though this is not required by \textit{grammatical} agreement; 
in the Gàidhlig sentence  
`Is ise am boireannach a chunnaic mi’ (that’s the woman I saw), the 
pronoun `ise’ (she) is feminine reflecting the natural gender of the  woman mentioned, 
but the article `am’ (the) is masculine in agreement with  the masculine noun `boireannach’ (a woman).
Although the study of linguistics teaches us to be wary of asking the \textit{purpose} of grammatical 
gender, we note that it can provide a measure of disambiguation in discourse, albeit inconsistently.
In French, `un livre' (masculine) is a book, whereas `une livre' (feminine) is a 
pound; in the English sentence `He went to see her mother' the varying pronouns tell us that three
people are involved, whereas the sentence `He went to see his mother' could refer to either two or three people.

In this research, we use a back translation technique to investigate
markers and drivers of gender bias using the popular DeepL
translation API. We chose DeepL for consistency with Treude and 
Hata's study \cite{treude_she_2023}, whose method we follow, and 
because a number of studies (including the work of Esperan{\c{c}}a-Rodier and Frankowski \cite{esperancca2021deepl}, 
and of Yulianto and Supriatnaningsih \cite{yulianto2021google}) have
demonstrated it to be one of the best translation APIs currently available.
Moreover, DeepL is widely used, has a
robust Python interface, and allows limited free use\footnote{See
https://www.deepl.com.}. 

We propose a new metric for gender
uncertainty in automatically translated text, arguing that this
measure avoids over-interpreting individual results, and
can generalize well across languages; we also find that the
main verb appearing in a given sentence is a significant driver of
gender uncertainty in translations. Further, we present some results
confirming the replicability of this approach. While we focus here on the
Software Engineering context, this approach could be generalized
to other domains of discourse.

While it is helpful to raise awareness of biases in language models, possibly partly hidden behind heuristic guardrails, this is still exploratory research, and further investigation is required to derive actionable steps to counterweigh bias in automatically translated (and otherwise automatically generated) texts. Treude and Hata have demonstrated the viability of the back-translation method with a minimal initial analysis, and we have significantly extended the approach to gain further insights, and take one further step towards eventual remediation. 

\section{Related Work}
\label{sec:related}

Globally, the workplace has suffered from bias or exclusion based
on gender, and the world economy would benefit significantly
from greater participation by women \cite{woetzel_power_2015}. 
Bias in software
development has been discussed in many articles, including those of Wang and Redmiles \cite{wang_implicit_2019}, Imtiaz \textit{et al} 
\cite{imtiaz2019investigating}, and Crick \textit{et al} \cite{crick_gender_2022}.
Garcia \textit{et al} discovered that female participants in Software Engineering
teams were more communicative and exhibited greater teamwork \cite{garcia_gender_2022},
while Terrell \textit{et al} noted that men's  contributions to open source projects
were accepted more readily than women's contributions\cite{terrell_gender_2017}.
Robillard has highlighted how gender bias leads to turnover in teams, which leads to loss of knowledge \cite{robillard2021turnover}.

Turning to bias in machine translations, Piazzolla \textit{et al} present 
a detailed study comparing popular
machine translation systems DeepL, Google Translate, and
ModernMT, finding that while DeepL better handles gender in
translation, all systems under-represent feminine forms \cite{piazzolla_good_2023}. De Vassimon Manela \textit{et al} quantify bias using a skew metric by examining stereotypical
and anti-stereotypical pronouns, and propose methods for
mitigation \cite{de_vassimon_manela_stereotype_2021}. 
Bordia and Bowman propose a metric to measure
bias, using a regularization procedure to encourage their machine
learning model to depend minimally on gender \cite{bordia_identifying_2019}. 
Tal \textit{et al} study the effect of model size on gender bias, and
conclude that while larger models make fewer gender errors, 
they also exhibit more bias \cite{tal_fewer_2022}.
Sun \textit{et al} present a good overview of 
approaches to identify and mitigate
bias, noting that `different applications may require different
metrics and there are trade-offs between different notions of
biases' \cite{sun_mitigating_2019}.

Back-translation has previously been used to investigate various characteristics of 
machine translation such as style transfer \cite{prabhumoye_style_2018}.
Treude and Hata propose adopting this technique to explore
bias in the translation of phrases in a Software Engineering context
\cite{treude_she_2023}. Their approach involves examining pronoun choice in
sentences back-translated from a language with gender-invariant
pronouns. The underlying assumption is that the appearance of particular pronouns
in differing contexts in the training data will influence the pronoun-selection
during the translation, perhaps disclosing a learned bias in the language model.
We should note however, that some terms appearing in the sentences used in these studies have a 
specific meaning in a Software Engineering context, but where these terms appeared in the 
training data of the language model, they may have been used with a more general meaning;
this may affect the association of a pronoun with its context in the model.

Treude and Hata discover that some tasks are more frequently
correlated with either `he’ or `she’ in the translation, and discuss
the relationship of task types and pronoun selection, noting the
more frequent appearance of particular pronouns with certain task types, and
present this as evidence of bias in the model. We build on and
extend this approach, as described in Section \ref{sec:method},
and we assess the replicability of this method in Section \ref{sec:repro}.

Sami \textit{et al} recently took a similar approach, but working with images instead of text \cite{mansour_sami_case_2023}; they examine the apparent
gender and ethnic diversity portrayed in images generated by Dall-E 2, using the same
set of 56 prompts, finding greater evidence of bias with images compared to
text-based studies.

In  Section \ref{sec:method}, we argue that the method of analysis used 
by Treude and Hata
and other authors rests on an unwarranted assumption that the appearance 
of certain pronouns is sufficient to indicate a bias, and propose an alternative.

\section{Research Method}
\label{sec:method}

We take a black-box approach to investigating bias in language models, by analysing the outputs from a translation/back-translation process.
We follow Treude and Hata's method \cite{treude_she_2023} by
automatically translating an English sentence containing `she' into
a language where third person pronouns do not reflect gender
(loosely called a `genderless language'), then translating back into
English. Since gender is not marked in the target language, the
DeepL API must select an English pronoun for the re-translation, 
which introduces an implication gender, revealing
biases in the underlying model. For example, translating the sentence
`As a software engineer, she performs support tasks.' into Finnish, the API returns
`Ohjelmistoinsinöörinä hän hoitaa tukitehtäviä.', where `hän' refers to
the software engineer without indicating gender; translating back to English gives
`As a software engineer, he takes care of support tasks.', revealing 
that the model renders the third person pronoun as `he', thereby eclipsing the
original feminine pronoun which appeared in the source sentence.

The 56 sentences used, each describing a Software Engineering
task, are taken from Treude and Hata's work \cite{treude_she_2023}, and ultimately derived from the
work of Masood \textit{et al} \cite{masood_like_2022}. The sentences
are presented in full in Treude and Hata's paper \cite{treude_she_2023}, and are also available in our Zenodo repository (see section \ref{available}). 

Each sentence was translated 100 times per run, randomizing the order each time, as 
the sequence of presentation may affect the translation. We see significant variation in
the pronouns selected across the 100 translations of each individual sentence;
these pronoun choices are the subject of our analysis.
As each of 56 sentences is translated 100 times, there are 5600 
back-translations in each dataset.

As all the original sentences use the pronoun `she', our analysis is based on quantifying the differing pronouns appearing in the re-translations. For each sentence, we tabulate occurrences of `he', `she', and other pronouns appearing in the back-translations. Exploring these patterns of variation may tell us something about biases in the model; for example, if certain co-located words, 
or descriptions of some particular activity, more often appear associated with a particular pronoun in the training data, we expect this to be reflected in the pronouns appearing in the back-translation of source sentences with corresponding words or activities. Our analysis aims to identify factors that affect pronoun variation, and to identify patterns using different `genderless' intermediate languages. Furthermore, we also investigate whether results are reproducible over time, using three time-lapsed data-sets for one of the languages (Finnish).

While it would be beneficial in future to use more sentences and different translation APIs, for the results of this pilot study to be comparable to earlier work, our data set is restricted to 5600 back-translations per language, as described. Given these restrictions on our data, we apply some statistical tests to establish the significance level of our results. 

\subsection{Research Objectives}
\label{sec:obj}

Our research objectives are as follow. First, we extend Treude and Hata’s methodology from one to five intermediate `genderless' languages, noting their comment that results from a single language may not be representative, and we compare the results across languages. Then, we seek a quantitative measure of pronoun selection without presupposing that the use of any particular pronoun is sufficient to signal a bias, avoiding the over-interpretation of individual words apparent in earlier research. 
Next, we look for higher-order patterns in our data, to identify whether any particular features may drive variation in pronoun selection, with the aim of moving the discussion beyond simply making assorted observations about individual isolated sentences. Lastly, we compare results from three time-lapsed datasets for the same language (Finnish), to investigate the replicability of results from the back-translation approach. These objectives are designed to inform our future research.

\subsection{Data Collection and Preparation}

Each sentence in the set was assigned an arbitrary ID taking
values 1, 2, …, 56, enabling consistent comparison over different
trials and different languages. For each trial, each sentence was translated 100 times in a
random sequence via the DeepL translation API, and a Python
script was used to extract the pronoun from the output and store
this against the sentence ID. The results were then analysed using the R
statistical programming language\footnote{See: https://www.r-project.org.}.

We derive the following datasets:

\begin{itemize}
\item FI  – results of back-translating via Finnish.
\item  INDO\footnote{We avoided using the label ID as this often indicates `identifier'.} – results of back-translating via Indonesian.
\item HU – results of back-translating via Hungarian.
\item TR – results of back-translating via Turkish.
\item ET – results of back-translating via Estonian.
\end{itemize}

When investigating the replicability of the back-translation method, we also use two
older datasets for Finnish, FI0 and FI1, which are introduced in Section \ref{sec:repro}.

\subsection{Data Availability}
\label{available}

Our data and code are available in a Zenodo repository at the following URL: https://zenodo.org/records/10522333.
Some charts and details of calculations are omitted here for brevity, but the relevant code is available in this repository. 

\section{Analysis and Results}
\label{sec:aar}

We start with some broad-brush observations, then present a more detailed analysis.
Counts for the pronouns appearing across all back-translated sentences for each language are
shown in Table \ref{tab:procount}.  Note that other possible pronouns
such as `she or he' and `one' are absent from the table as they never appeared in any output.

\begin{table}[htbp]
\centering
\caption{Pronoun count across all datasets}
\begin{tabular}{|c|c|c|c|c|c|} 
\hline
 & FI & HU & INDO & TR & ET \\
\hline
(none) & 6 & 4215 & 0 & 4353 & 32 \\
\hline
he & 4540 & 325 & 5595 & 38 & 3030 \\
\hline
he or she & 234 & 33 & 0 & 71 & 331 \\
\hline
he/she & 813 & 309 & 4 & 383 & 2021 \\
\hline
it & 0 & 0 & 0 & 1 & 0 \\
\hline
she & 3 & 6 & 1 & 1 & 186 \\
\hline
they & 2 & 0 & 0 & 0 & 0 \\
\hline
you & 2 & 712 & 0 & 753 & 0 \\
\hline
\end{tabular}
\label{tab:procount}
\end{table}

We observe that Finnish and Estonian, related languages spoken in the same geographical region,
have a similar profile. Both use the pronoun `he' extensively in translations, and both rarely
render a back-translated sentence with no pronoun. `He or she' and `he/she' appear with moderate
frequency, though more frequently in Estonian, and Estonian also shows the greatest 
use of `she' overall with 186 occurrences. These two languages show the greatest variation in pronoun 
selection over all translations. 

Turkish and Hungarian show a broadly similar pattern to each other, with moderate use of
`he or she' and `he/she', but greater use of `you' which barely appears with the other languages.
Noticeably, Turkish and Hungarian both render many back-translations with no pronoun at all, 
which is possibly an artefact of the sentence structure in these languages. 

Finally, Indonesian
shows a different pattern, using `he' almost exclusively, with only five instances of any other
pronoun appearing at all. While this may tell us something about the Indonesian language model
as a whole, this dataset is little used in our following analysis, owing to this lack of variation.

Overall, we might conclude that the Estonian back-translations are in some sense least biased, as
they make the greatest use of `he or she', make by far the greatest use of `he/she', and show
greater variation across  all sentences;
however, the pronoun `he' is nonetheless much used in this dataset.

\begin{figure}[htbp]
\centering
\includegraphics[scale=0.45]{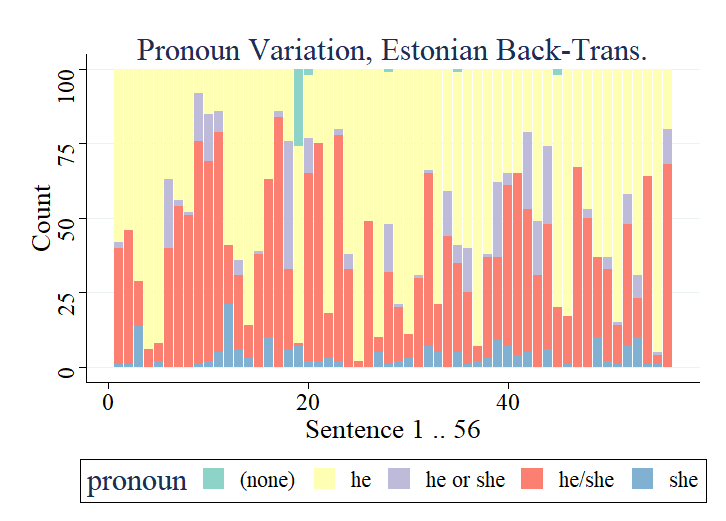}
\caption{Pronoun distribution across sentences for Estonian}
\label{fig:ET-pro}
\end{figure}

To better illustrate how the varying pronouns are distributed across the individual sentences, rather than in the dataset as a whole,
Fig. \ref{fig:ET-pro} shows the frequency with which pronouns appear for each sentence in the
Estonian back-translation. (Charts for the other languages are omitted here for brevity, but
can be generated using the code in our Zenodo repository). 
Each bar represents a sentence, with the fill showing the different pronouns used across the 100 translations of that sentence; the numbering of the sentences follows their ordering in the earlier studies, and is not significant. 
Here we see the wide use of `he' (pale yellow background colour), and considerable use of `he/she' (red) for some sentences. 
The occasional use of `she' is marked (in blue) at the base of some bars.

In summary, we can see that the languages used fall into three loose groups:

\begin{itemize}
    \item Group 1: Finnish and Estonian -- few missing pronouns, frequent use of `he' and moderate use of `he/she' and `he or she', little use of `you'.
    \item Group 2: Hungarian and Turkish -- many missing pronouns, comparable use of `he' and `he/she' or `he or she', greater use of `you'.
    \item Group 3: Indonesian -- almost exclusive use of `he'.
\end{itemize}

\subsection{Metrics}
\label{metrics}

To interpret the results of these experiments, we need to consider the pronouns
that appear in the translations. 
Seven different pronouns (including \textit{none} as a choice)
appear across all the translations.
Our initial discussion focuses on `he', `she', and
combinations such as `he or she', as other choices such as `you', `they', or 
`one' appear less frequently in the translations, and moreover are
free of any implied gender.

In English, `he' refers to male gender, and
`she' to female. However, there is a long-standing usage where `he' may refer
to any or to unknown gender; the Cambridge dictionary
says\footnote{See: https://dictionary.cambridge.org/dictionary/english/he.} it 
can be `used to refer to a person whose gender is not known or not important in that situation'.
This is more common in traditional usage;
proverbs such as `He who hesitates is lost' were not intended to refer only to males. 
Nowadays, this epicene use of `he’ is becoming increasingly uncomfortable;
Wiktionary states `… since the mid-20th century generic
usage has sometimes been considered sexist and limiting… In
place of generic he, writers and speakers may use he or she,
alternate he and she as the indefinite person, use the singular they,
or rephrase sentences to use plural they’ \cite{noauthor_wiktionary_2023}.

In earlier research, other authors have presented usage of `he' as evidence of bias in the
translation. However, this is based on the premise that `he' indicates only
male gender, and overlooks the epicene sense of `he’.
We suspect that epicene `he' must be embedded in language models,
especially where these have been trained on texts following older writing conventions.
Arguably, this male-only interpretation brings a level of bias even to the analysis.

Another issue  is that 
Treude and Hata's method of calculating the proportion of male referents is unbalanced, 
including `he/she' and `he or she' in female totals but not in male.
Sami \textit{et al} discovered this and corrected the calculation in their 
image-based study \cite{mansour_sami_case_2023};
however, their subsequent analysis still rests on the presumption that the pronoun `he' can refer only to males.

To summarize, we may understand pronouns in translations as follows:

\begin{itemize}
    \item She -- indicates female gender. This is the only pronoun indicating one particular gender unambiguously.
    \item He -- indicates male gender, but may also indicate both genders or indeterminate gender (epicene `he').
    \item They, you, one -- does not indicate gender.
    \item He/she, he or she -- indicates both male and female gender. Usage usually signals an awareness of implied gender
    and an intention to avoid this.
\end{itemize}

Given the huge datasets used in training large language models, we must suppose
that instances of epicene `he' were present in training corpora and that
the appearance of `he' cannot therefore be interpreted only as indicating male gender only;
indeed, the only pronoun that clearly indicates a specific gender is `she', which suggests
we may get more insight into bias in the models by counting appearances of `she' rather
than of `he'.

Furthermore, there is also a presumption that `he or she', or its variant `he/she', represents
an unbiased pronoun choice. But contrasting `he' with `she' obviates the epicene
sense of `he', and is taken to mean specifically male or female; this excludes non-binary identities.
Furthermore, does writing `he or she' instead of `she or he' also imply a bias?
The very rare usage of `she or he' (which does not appear in any translations) seems
to indicate little awareness of the potential connotation of ordering the pronouns with
`he' in the first position. 
Our conclusion is that picking out individual pronouns as indicators of bias
is fraught with presuppositions and subject to individual interpretations.
We need to try a different approach. 

We propose
therefore to examine variability of the pronoun selection. Where a
sentence is repeatedly translated with `he', whether we take that to
be inclusive (epicene) or exclusive (masculine), it is certainly
consistent; where varying pronouns appear in repeated
translations of the same sentence, we argue that the language
model displays greater uncertainty, we may even say hesitancy,
in implying a particular gender. While we cannot impute a specific bias in
this way, we can identify sentences whose translations
demonstrate greater sensitivity to the pronoun selection. We see a
parallel to human usage, where an English speaker saying `he or she’, or
(increasingly) `they’ when referring to an indeterminate person
thereby signals their awareness of implying the gender of
the person spoken of, and their intention to avoid this. We believe this new approach can give some
valuable insights.

To quantify pronoun variation, we use a scaled version of the `coefficient of
unalikeability' (UC), introduced by Perry and Kader \cite{perry_variation_2005}, and later
expanded by the same authors \cite{kader_variability_2007}. 
Perry and Kader define unalikeability as a measure of how often observations differ,
and note that the measure focuses on `how often the observations differ, not how much.'
The unalikeability coefficient gives a measure
of variation in a categorical variable, analogously to 
how the standard deviation measures variability in a continuous variable.
The UC takes values from zero to one, representing a scale from no variability (0) to maximum variability (1). 

We consider the repeated
translations of a given sentence as 100 observations of a
categorical variable with 7 levels, and calculate the UC
accordingly. However, in 100 translations, to achieve UC = 1 (maximum variation)
would require 100 different pronouns. With only 7 pronouns used,
the maximum possible UC, which we call UC$_{max}$, is 0.866. We therefore
normalize the UC value to give an `adjusted UC value', UCA,
which more intuitively extends across the range of possible
variation when selecting a pronoun. A value of UCA = 1 would 
indicate that all 7 pronouns are used equally in a given translation.
Our measures are defined in equations (1) and (2)
below. The definition of UC given is taken from Perry and Kader \cite{perry_variation_2005}.

\begin{samepage}
\begin{equation}    
UC = \sum_{i \neq j} \frac{c(x_{i} , x_{j})} {(n^2 - n)}
\end{equation}
\nopagebreak
where $c(x_{i} , x_{j}) = 1$ if $x_{i} \neq x_{j}$, 
and $c(x_{i} , x_{j}) = 0$ otherwise.
\nopagebreak
\begin{equation}
UCA = \frac{UC} {UC_{max}}   
\end{equation}
\end{samepage}

Here, $x_{i}$ and $x_{j}$ are the values of  the categorical variable (pronoun choice) compared pairwise for each of 
the 100 translations of the same sentence in each dataset; so $n = 100$. UC$_{max}$ is the maximum
possible value of UC (0.866 for our datasets) given the 7 pronouns appearing in
all back-translations. The calculation showing UC$_{max} = 0.866$ is given in
our Zenodo repository. 

Looking at the overall UCA for each dataset, shown in Table \ref{tab:uca-dist},
we see that with the exception of Indonesian, where  nearly all pronouns are rendered as `he',
the distribution of UCA values appears similar across the other four
languages. The mean value and the position of the first and third quartiles are broadly comparable,
though somewhat higher for Estonian which shows greater variability overall.
Likewise, the standard deviation of UCA values is similar across these four datasets.
This suggests that the UCA metric has potential to generalize well across languages.

\begin{table}
\centering
\caption{Distribution of UCA across sentences in each dataset}

\begin{tabular}{|l |l |l |l |l |l|} \hline  
 & FI & HU & INDO & TR & ET \\ \hline 
Min. & 0.000 & 0.046 & 0.000 & 0.000 & 0.046 \\ \hline 
1st Qu. & 0.112 & 0.253 & 0.000 & 0.190 & 0.409 \\ \hline 
Median & 0.265 & 0.406 & 0.000 & 0.393 & 0.583 \\ \hline 
Mean & 0.303 & 0.404 & 0.002 & 0.379 & 0.523 \\ \hline 
3rd Qu. & 0.498 & 0.532 & 0.000 & 0.558 & 0.639 \\ \hline 
Max.  & 0.730 & 0.859 & 0.046 & 0.888 & 0.824 \\ \hline 
Std. Dev. & 0.217 & 0.203 & 0.009 & 0.225 & 0.189 \\ \hline

\end{tabular}
\label{tab:uca-dist}
\end{table}

A higher value of UCA will highlight a sentence where the pronoun tends to vary 
on back-translation, and a lower value will indicate a sentence where the same pronoun
-- regardless of whether it is `he', `she', or something else -- tends to be used.

As expected, different sentences show different degrees of pronoun 
variation according to the language. For example, the first sentence `As
a software engineer, she identifies constraints' shows low variation
back-translating from Finnish (UCA = 0.07) but high variation
back-translating from Estonian (UCA = 0.60). We would expect that different
data corpora used in training would lead to different areas of bias
in the various language models, and that we may find patterns
in the variation across sentences for any given language. 
Each language merits further individual study.

However, it may also be interesting to examine
which sentences have high or low variability in pronoun selection across all languages.
Table \ref{tab:hi-lo_uca} shows those sentences where the mean UCA across languages
(excluding Indonesian) is either in the fourth quartile averaged over the dataset (high
variability), or is in the first quartile (low variability).
In this table, the prefix `As a software engineer ...' is omitted for formatting. 

\begin{table}
\centering
\caption{Sentences with high and low UCA across datasets}
\begin{tabular}{|l |l |l |l |l|} \hline  
sentence & FI & HU & TR & ET \\ \hline 
she performs user training. & 0.497 & 0.796 & 0.570 & 0.496 \\ \hline 
she asks coworkers. & 0.606 & 0.439 & 0.888 & 0.667 \\ \hline 
she stores design versions. & 0.554 & 0.233 & 0.689 & 0.794 \\ \hline 
she submits changes. & 0.662 & 0.402 & 0.656 & 0.662 \\ \hline 
she manages development branches. & 0.477 & 0.474 & 0.626 & 0.764 \\ \hline 
she releases code versions. & 0.297 & 0.635 & 0.684 & 0.713 \\ \hline 
she has meetings. & 0.267 & 0.686 & 0.701 & 0.616 \\ \hline 
she performs infrastructure setup. & 0.662 & 0.530 & 0.485 & 0.564 \\ \hline \hline
she restructures code. & 0.068 & 0.443 & 0.254 & 0.132 \\ \hline 
she reads changes. & 0.173 & 0.277 & 0.090 & 0.406 \\ \hline 
she reads artifacts. & 0.068 & 0.112 & 0.091 & 0.653 \\ \hline 
she writes artifacts. & 0.023 & 0.175 & 0.152 & 0.582 \\ \hline
\end{tabular}
\label{tab:hi-lo_uca}
\end{table}

We can see that the high UCA sentences include two tasks that require `performing',
two that involve asking/meeting others, and several that suggest administrative activities.
On the other hand, the smaller set of low UCA tasks centres on the seemingly more definitive 
activities of reading, writing, and restructuring.

It is interesting that the `perform' sentences all have a similar level of 
variability (as shown in Section \ref{sec:verb}) despite characterizing the largest
group of tasks. Furthermore, these results show a notable concordance with 
Treude and Hata's more \textit{ad hoc} analysis \cite{treude_she_2023}, 
where they report that `perform infrastructure setup, perform support
tasks were associated with ``he'' in the minority of cases', while tasks including `restructure code, write artifacts'
were `associated with “he” in at least 99 out of 100 runs'.

The real world implication is that in generated texts, certain tasks or types of tasks are 
presented with several pronouns varying over different occurrences, whereas other tasks are regularly
shown with with little or no variation. This has the potential of subtly reinforcing gender stereotypes over repeated exposure to such generated texts.

\subsection{Sentence Classification}
\label{sec:verb}

To gain a broader understanding of whether different types of
sentences (reflecting different types of activities) can affect the
UCA of a translation, we consider classifying the 56 sentences
used. The task classification of Masood \textit{et al} \cite{masood_like_2022} is useful for the
Software Engineering domain, but although we focus on this area,
we seek an approach that may be more generalizable. 

We considered classifying sentences based on sentiment scores
of the words they contained; however, 
the Software Engineering vocabulary used is poorly represented in
standard sentiment dictionaries such as NRC and AFINN, making it 
difficult to label the sentences.
(D’Andrea \textit{et al} give a good overview of sentiment analysis 
tools \cite{dandrea_approaches_2015}). 

Another approach, also discussed by Treude and Hata \cite{treude_she_2023}, is to
consider the `pink tasks’ identified by Garcia \textit{et al} \cite{garcia_gender_2022}, which are said to be associated with `perceived feminine competencies'.
However, this offers only a binary classification of tasks, and moreover
seems somewhat subjectively defined.

Seeking a better approach, we note that 37 distinct verbs appear
in the 56 sentences, and posit that verbs carry much of the
semantics of the sentence. We expect that writing an email and
writing documentation have a common theme. Therefore, we
calculate UCA for groups of sentences containing the same verb.
We limit this analysis to the FI, HU, TR and ET datasets, 
as the Indonesian dataset (INDO) shows almost no pronoun variation.

For robustness, we restrict consideration to verbs that appear at
least twice across the sentences, which leaves 9 of the 37 distinct 
verbs\footnote{Sentence 26 `she reads/reviews code' raised an issue, 
as Masood \textit{et al} did not commit to a single verb here; 
we used `read' in this case, as `review' does not appear elsewhere.}.
Table \ref{tab:uca-verb} shows the averaged
UCA values for each verb. The `count' column shows how many
times each verb appears in different sentences, and 
each language column shows UCA$_{mean}$, the mean value of UCA for all sentences
containing that verb.

\begin{table}
\centering
\caption{ Mean value of UCA for sentences grouped by their verb}
\begin{tabular}{|l |l |l |l |l |l|} \hline    
verb & count & FI & HU & TR & ET \\ \hline 
browse & 4 & 0.222 & 0.531 & 0.629 & 0.531 \\ \hline 
edit & 2 & 0.351 & 0.382 & 0.502 & 0.185 \\ \hline 
fix & 2 & 0.319 & 0.555 & 0.419 & 0.377 \\ \hline 
perform & 5 & 0.615 & 0.389 & 0.282 & 0.531 \\ \hline 
produce & 3 & 0.098 & 0.444 & 0.358 & 0.551 \\ \hline 
provide & 4 & 0.115 & 0.453 & 0.496 & 0.462 \\ \hline 
read & 3 & 0.214 & 0.160 & 0.083 & 0.547 \\ \hline 
submit & 2 & 0.446 & 0.268 & 0.404 & 0.648 \\ \hline 
write & 3 & 0.111 & 0.152 & 0.219 & 0.592 \\ \hline

\end{tabular}
\label{tab:uca-verb}
\end{table}

Hungarian and Turkish agree that `browsing' is one of the most gender-uncertain
activities (showing greatest pronoun variation), and that `reading' and `writing' are the most
certain activities (showing least pronoun variation), while Finnish and Estonian
agree that `submitting' is the most gender-uncertain activity. Otherwise,
each language shows a distinctive pattern of UCA for each verb, perhaps 
reflecting underlying variation in the data used for training the language model.
More generally, we find that that
sentences grouped by verb have distinctive values of UCA within each language,
implying the verb is a driver of gender uncertainty in each
underlying model.

\begin{figure}[htbp]
\centering
\includegraphics[scale=0.42]{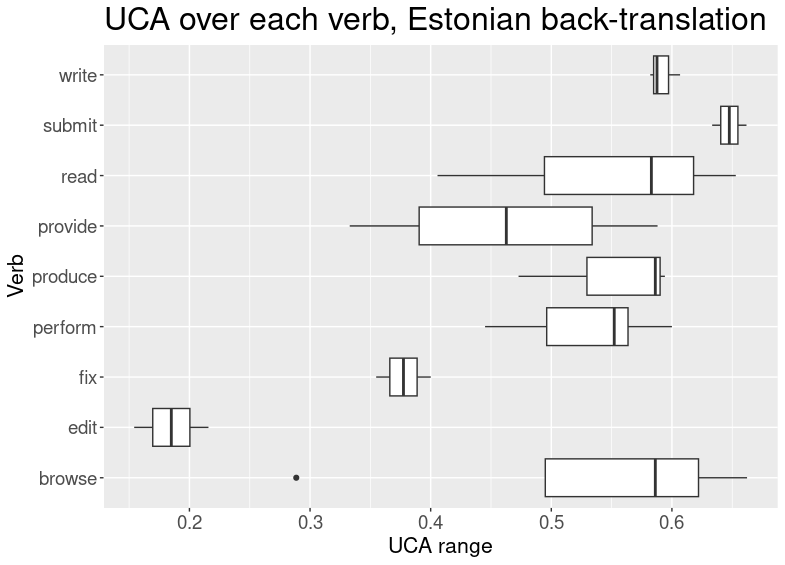}
\caption{UCA per verb, Estonian back-translation}
\label{fig:uca_per_verb_et}
\end{figure}

In Fig. \ref{fig:uca_per_verb_et}, we see the distribution of UCA values over
different sentences using the same verb, for the Estonian dataset. (Charts for the other languages
are omitted for brevity, but relevant code is available in our Zenodo repository). 
Here we
note the lower overall UCA of the verb `edit', and we see that some verbs have a 
tighter range of uncertainty than others. This may simply reflect how many sentences
use each verb; however, the most frequent verb `perform', which occurs
in five sentences, has a relatively tight range of UCA values, indicating 
consistent levels of pronoun variation in different sentences using this verb.

Although these results are promising, and suggest that the main
verb in a sentence is a driver of its UCA, each verb occurs only a few times,
and some occur in more sentences than others. This is unsurprising, as the
56 sentences were not designed for this kind of analysis; but we may still have
enough evidence to support the finding.

For each dataset, we therefore performed an ANOVA (analysis of variation) test
to establish the statistical significance of the variation between different
sentences with the same verb. We defined nine groups,
each containing all the sentences with the same verb, and compare
the mean UCA values of each group. The results are shown in
Table \ref{tab:anova}. (The code can be found in our Zenodo 
repository).

\begin{table}
\centering
\caption{ANOVA results showing the main verb is a driver of gender variation}

\begin{tabular}{|l |l |l|} \hline  
 & F-value & P-value \\ \hline 
FI & 7.516 & 0.0002 \\ \hline 
HU & 2.188 & 0.0771 \\ \hline 
TR & 3.102 & 0.0204 \\ \hline 
ET & 4.267 & 0.0045 \\ \hline

\end{tabular}
\label{tab:anova}
\end{table}

The F-value shows the ratio of between-group variation to within-group variation; 
higher values indicate that the groups for each
verb are more internally consistent and more mutually distinct. 
The p-value shows the statistical significance for each group. 
The test achieves statistical significance at the level
$p \leq 0.05$, except for Hungarian which falls just short at $p = 0.08$.
These results seem promising, given the small number of observations used, and suggest that classifying by verb reliably reveals some higher-order patterns in the way different sentences
are translated.

For further insight, we performed a Tukey range test \cite{tukey1949comparing} 
for each language, and found a statistically significant difference at the 95\% confidence level
for several individual verb pairings for both Finnish and Estonian, and one for Turkish.
For Finnish, the pairs were
perform/browse (p=0.004), produce/perform (p=0.0005), provide/perform (p=0.0003), read/perform (p=0.008), write/perform (p=0.0007);
for Estonian they were edit/browse (p=0.01), perform/edit  (p=0.01), produce/edit  (p=0.01), read/edit  (p=0.02),
and submit/edit  (p=0.004); and for Turkish, read/browse (p=0.01). 

Despite the small number of observations, we have good evidence that the main verb in a sentence drives the  variation of pronouns in the translation, especially where the language displays more variability overall (Group 1). This is a promising result and merits more detailed investigation.

\subsection{Reproducibility}
\label{sec:repro}

Treude and Hata noted that they had not established reproducibility of results from the back-translation
method; we can broach this for Finnish, as we have three datasets available for time-separated
replications of the same experiment. These datasets are:

\begin{enumerate}
    \item FI0 -- Treude and Hata's dataset, made available by these authors; 
    the files in their Zenodo repository are dated March 2023.
    \item FI1 -- Data from our early replication of the Finnish study in June 2023.
    \item FI -- The main Finnish dataset discussed here, created in October 2023.
\end{enumerate}

As each of these Finnish datasets was produced in the same way, they allow us to compare
consistency of results over time, with approximately three then four months' respective 
separation between them. Table \ref{tab:fin3-procount} shows the overall distribution of pronouns 
in each dataset.

\begin{table}
\centering
\caption{Pronoun count for three time-separated Finnish datasets}
\begin{tabular}{|l |l |l |l|} \hline  
 & FI0 & FI1 & FI \\ \hline 
(none) & 21 & 14 & 6 \\ \hline 
he & 4490 & 4491 & 4540 \\ \hline 
he or she & 0 & 0 & 234 \\ \hline 
he/she & 842 & 849 & 813 \\ \hline 
she & 240 & 244 & 3 \\ \hline 
they & 1 & 0 & 2 \\ \hline 
you & 6 & 2 & 2 \\ \hline

\end{tabular}
\label{tab:fin3-procount}
\end{table}

Here we see a broadly similar usage of `he' and `he/she' across all three datasets,
with minimal use of `they' and `you' in all cases. (There is one instance of `they' in
Treude and Hata's dataset, though it is not discussed in their paper).
For these pronouns, we see a good level of consistency across all three datasets.

However, the count of `she', comparable at around 240 occurrences in the first two datasets, drops 
dramatically to only 3 instances in the most recent dataset. Furthermore, the latter FI
dataset has 234 instances of `he or she', which 
did not appear at all in the first two datasets. Clearly there has been a change in the behaviour
of the DeepL API between the creation of the second and third datasets. As the most recent dataset has
234 instances of `he or she', while the first two had 240 and 244 instances respectively of 
`she', it appears that where previously the API rendered `she', it now often renders `he or she'.
Arguably this is a reduction in bias, as `she' refers only to female gender,
whereas `he or she' includes both male and female. 

Eyeballing the data sentence by sentence, it is apparent that not every instance of `she' in
the older two datasets has been uniformly replaced by `he or she', but the results have largely shifted in
this direction.  This could be due to further training in the DeepL language model,
but it seems more likely due to the recent addition of a heuristic to increase
usage of `he or she'. We note that the stand-out sentence `As a software engineer, she elicits requirements', which used `she' 43 times in Treude and Hata's dataset, 
giving their paper its
title, no longer returns a single instance of `she' in the most recent dataset.

In order to establish if our UCA metric is stable over time, and especially
in the face of the shift in pronoun usage between the second and third Finnish datasets, we examine 
the overall distribution of UCA values in each dataset, shown in Table \ref{tab:repro-uca}.
Here we see a good level of consistency across all three time-lapsed datasets.

\begin{table}[b]
\centering
\caption{Distribution of UCA values over time-lapsed Finnish datasets}

\begin{tabular}{|l |l |l |l |l |l |l|} \hline   
 & Min. & 1st Qu. & Median & Mean & 3rd Qu. & Max. \\ \hline 
FI0 & 0.0000 & 0.1330 & 0.2770 & 0.3135 & 0.4855 & 0.7225 \\ \hline 
FI1 & 0.0000 & 0.1690 & 0.2724 & 0.3149 & 0.4807 & 0.7136 \\ \hline 
FI & 0.0000 & 0.1120 & 0.2652 & 0.3026 & 0.4975 & 0.7295 \\ \hline

\end{tabular}
\label{tab:repro-uca}
\end{table}

\begin{figure}[htbp]
\centering
\includegraphics[scale=0.43]{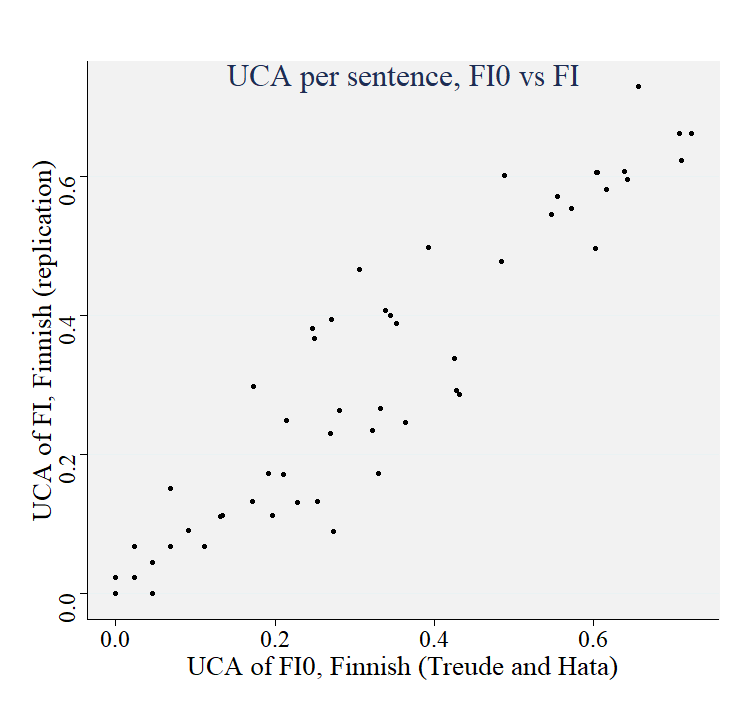}
\caption{UCA per sentence, FI0 versus FI datasets}
\label{fig:uca-scatter}
\end{figure}

To establish whether each individual sentence also maintains a consistent UCA value over time, 
we plot UCA values pairwise for each of the 56 sentences between the two Finnish datasets most
separated in time, FI0 and FI. Fig. \ref{fig:uca-scatter} shows a strong correlation, with the same sentence broadly  returning a similar UCA value in each dataset despite the shift in pronoun usage.

Calculating Pearson's correlation co-efficient between the FI dataset and each of the previous
two Finnish datasets, FI0 and FI1, we see a high degree of correlation, with values 0.936 and
0.953 respectively. However, calculating the correlation for how often `she' appears in
translations of each sentence, we get a low level of correlation with values 
0.087 and 0.055 respectively. This shows that while the use of the pronoun `she' has changed substantially
in the API between the second and third experiments, and does not provide a reproducible measure,
the UCA metric has remained stable. It should be noted that while the count of `he' also shows a good
level of consistency, this is by far the commonest pronoun overall, and as we noted earlier,
it is unclear if it should be interpreted in a masculine or an epicene sense.

\section{Discussion}
\label{sec:disc}

We have extended the work described by Treude and Hata \cite{treude_she_2023} in the following ways: 

\begin{enumerate}
    \item We extend the approach from one to five intermediate languages,
            using all the languages available in the DeepL API with suitably invariant third-person pronouns;
           \textit{viz}. Finnish, Estonian, Turkish, Indonesian, and Hungarian;
    \item we propose a novel metric for assessing uncertainty of gender in the translation;
    \item we investigate sentence features which drive implications of gender in translations; and
    \item we compare three time-lapsed datasets for Finnish, to establish the replicability of the approach.

\end{enumerate}

Our main findings are as follow.

\textbf{Finding 1}: We see that using several target languages results 
in different patterns of pronoun usage, showing the need to work with multiple
languages to achieve generalizable results. Our analysis suggests that the five languages
used fall into three loose groupings, with similar distribution of pronoun use within
each group, and with broadly similar distribution of UCA value across two of these groups.

\textbf{Finding 2}: We note that singular `they', which is increasingly 
common in modern usage \cite{lascotte_singular_2016}, 
is vanishingly rare in all back-translations. This suggest that the language 
models have been trained on data lagging behind current
usage, implying that we need to consider the epicene sense of `he' in our analysis. For this reason,
we take care not to over-interpret the appearance of 
any given pronoun, noting that only `she' clearly indicates a particular gender.
We therefore propose the adjusted unalikeability coefficient as a suitable metric to investigate our data.
This measure shows where a language model displays greater variation in selection of pronoun,
which we take as a surrogate for sensitivity to implication of gender;
the UCA values  appear  robust across languages and across replications, and we
believe it will be useful in future research.

\textbf{Finding 3}: We see good evidence that the verb appearing in a
phrase is a driver of gender uncertainty, with significant
difference in UCA for sentences grouped by their main verb.
These higher-level patterns of gender uncertainty in the
translations require further investigation, to enable analysis of
back-translation data to rise
above the level of over-interpreting individual observations.

\textbf{Finding 4}: We have observed a change in the behaviour of the DeepL
translation API during the course of the study, perhaps caused by addition
of a heuristic to address gender bias. We found that while counts of the 
pronoun `she' in each sentence were entirely disrupted by this update,
the pronoun variation measured by UCA still showed a good level of per-sentence
correlation across the API change, suggesting that it is a robust metric
for studying bias, and that the back-translation method gives reproducible
results overall.

These findings will inform our future research, as outlined in
Section \ref{sec:future}.

\subsection{Limitations of this Study}

Although this study is exploratory in nature, not yet seeking to confirm specific
research questions requiring a more formal analysis of threats to validity \cite{wohlin_challenges_2021},
some brief remarks on limitations of this research follow. Further limitations are implied by the ideas for further work outlined in Section \ref{sec:future}.

While we have extended earlier work by investigating more languages, 
we need also to compare results from more translation APIs, 
as DeepL may be in some ways atypical. At present, we cannot say if the loose grouping of languages we observed based on their pronoun profile would be consistent across different language models. Moreover, the 
5600 translations created per dataset still constitute a small-scale experiment, 
and more data are required to gain insight into potential of the method, especially as we have noted a 
change in the behaviour of the DeepL translation API during our study. While our results give a strong indication that the 
main verb is a significant driver of gender variation in translations, the datasets are not well structured to support 
this analysis, as only nine verbs appear twice or more, and sentences are not balanced for verb usage. 
This can be addressed by a future experiment designed specifically to test the robustness of this correlation between verb and pronoun variation.

\subsection{Future Work}
\label{sec:future}

This exploratory study was intended to identify lines for future research, 
and our results suggest several interesting continuations.

Noting one pattern of broadly similar pronoun usage for
Finnish and Estonian, and another pattern for Turkish and Hungarian,
we suspect that language models may fall into loose groups,
and languages within each group should be compared in greater detail.

Although the data from our Indonesian translations were not suitable
for analysis here, it may be that longer sentences or text fragments would 
produce more useful results for this language, and this should be investigated.
Given the almost exclusive use of `he' in Indonesian back-translations,
it may also be interesting to investigate other markers of bias in the
Indonesian language model.

In an earlier trial, not reported here, where we back-translated from Finnish
the same 56 sentences but omitting the prefix `As a Software Engineer',
we obtained short agrammatical and fragmentary outputs, often lacking any pronoun,
as in the Indonesian data. This may indicate that a minimum
length of text is required for each  language to get a reliable result. 
In general, we expect  that use of longer
phrases for translation will capture more context-dependent bias in 
the language model, and give further insights,
both for Software Engineering and for other domains.

The verb approach shows promise in
identifying higher-level patterns of implied gender uncertainty,
but this requires more corroboration. Here we used the 56 sentences from earlier 
studies for comparability, but we plan an experiment using different sentences
where each verb appears the same number of times, in a greater number of examples, 
and with varying contexts and collocated words for each verb.
This will allow clearer and more statistically significant results.

It would also be interesting to investigate correlation of pronoun selection with
sentiment labels for words appearing in the sentences, 
but this would need to be done either in a more general context,
or using a custom sentiment dictionary that covers Software Engineering
terminology. 

Finally, we note that it may be useful to entirely invert the
methodology of this study. Instead of translating multiple
sentences with a fixed contextual prefix such as `As a Software Engineer', 
rather we could translate the
same, rather general sentences with differing contexts. For
example, simple phrases such as `she writes’ or `she helps others’
could be prefixed by `As a software architect’, `As a team-leader’,
\textit{etc}. This approach should further illuminate the role of context in
the source text for eliciting implications of gender, and could be
further broadened using contexts such as `At her work, …’, `As a
law-maker, …’, `As a parent, …’, \textit{etc}.

\section{Conclusion}
\label{sec:conc}

Extending Treude and Hata’s back-translation approach shows it
has great potential, with much still to be realized. Our approach of
identifying gender variation in the translations gives a new
perspective on where bias may occur.

We argue that it is hard to say what features of text constitute a bias,
and that we may not even know what non-biased text should look like.
For example, according to data from the Office for
National Statistics, in the UK 97\% of veterinary nurses are female
\cite{uk_government_employment_nodate} (retrieved 2023); 
should `As a veterinary nurse …’ lead to selection of `she' in
97/100 translations? Or should `he' and `she' vary randomly, each appearing 
in 50\% of the sentences? Using the UCA metric allows us
to probe biases without having to make any questionable 
assumptions about what constitutes a biased formulation in the first place. 

To conclude with a contrarian view, we might argue that our
language models work just fine, accurately reflecting the biases of our
society; if change is desirable, society must change first,
and the language models will follow. However, sensitivities are changing towards use of
language, and models trained on corpora reflecting older forms of
usage might entrench biases that re-appear in newly generated
text, inadvertently perpetuating those biases; for example, we
noted earlier the scarcity of singular `they’ in translations despite
its widespread colloquial use. Further investigation of how bias in existing
models is manifested will help address such broader issues.

\section*{Acknowledgment}

We are pleased to thank Christos Chrysoulas and Tess Watt, for their comments; our anonymous SANER 2024 referees, for their feedback; and Lisa Smart, for proof-reading.

\bibliographystyle{IEEEtran}
\bibliography{llmbias}
\vspace{12pt}

\end{document}